%% file: main.tex
\documentclass{article} %
\usepackage[accepted]{icml2021}
\usepackage[utf8]{inputenc}
\usepackage{amsmath}
\usepackage{amssymb}
\usepackage{array}
\usepackage{xcolor} 
\definecolor{mydarkblue}{rgb}{0,0.08,0.45}
\usepackage[colorlinks,citecolor=mydarkblue,urlcolor=mydarkblue,linkcolor=mydarkblue,filecolor=mydarkblue]{hyperref}
\usepackage{url}
\usepackage{relsize}

\usepackage{multirow}
\usepackage{booktabs}
\usepackage{caption}
\usepackage{subcaption}
\usepackage{natbib}
\usepackage{graphicx}
\usepackage{makecell}
\usepackage{tabu}
\usepackage[hang,flushmargin]{footmisc} %

\setcitestyle{numbers,square}

\input{defs.tex}

\usepackage[textsize=scriptsize,textwidth=2.2cm]{todonotes}
\usepackage{etoolbox}
\newtoggle{showtodos}
 \usepackage[compact]{titlesec}
 \titlespacing{\section}{0pt}{1ex}{0ex}
 \titlespacing{\subsection}{0pt}{1ex}{0ex}

\begin{document}
\cfoot{\thepage}
\setlength{\footskip}{3em}

\twocolumn[
\icmltitle{A Simple Fix to Mahalanobis Distance for Improving Near-OOD Detection}

\icmlsetsymbol{equal}{*}

\begin{icmlauthorlist}
\icmlauthor{Jie Ren}{brain}
\icmlauthor{Stanislav Fort}{equal,stan}
\icmlauthor{Jeremiah Liu}{equal,brain,harvard}
\icmlauthor{Abhijit Guha Roy}{health}
\icmlauthor{Shreyas Padhy}{brain}
\icmlauthor{Balaji Lakshminarayanan}{brain}
\end{icmlauthorlist}

\icmlaffiliation{stan}{Stanford University}
\icmlaffiliation{harvard}{Harvard University}
\icmlaffiliation{brain}{Google Research}%
\icmlaffiliation{health}{Google Health}

\icmlcorrespondingauthor{Jie Ren}{jjren@google.com}
\icmlcorrespondingauthor{Balaji Lakshminarayanan}{balajiln@google.com}

\icmlkeywords{Machine Learning, ICML}

\vskip 0.3in
]

\printAffiliationsAndNotice{\icmlEqualContribution} %

 \begin{abstract}
 Mahalanobis distance (MD) is a simple and popular post-processing method for detecting out-of-distribution (OOD) inputs in neural networks. We analyze its failure modes for near-OOD detection and propose a simple fix called \textit{relative Mahalanobis distance} (RMD) which improves performance and is more robust to hyperparameter choice. On a wide selection of challenging vision, language, and biology OOD benchmarks (CIFAR-100 vs CIFAR-10, CLINC OOD intent detection, Genomics OOD), we show that RMD meaningfully improves upon MD performance (by up to $15\%$ AUROC on genomics OOD).
\end{abstract}

\section{Introduction}
Out-of-distribution (OOD) detection is critical for deploying machine learning models in safety critical applications \citep{amodei2016concrete}. A lot of progress has been made in improving OOD detection by training complicated generative models \cite{bishop1994novelty,nalisnick2019detecting, ren2019likelihood, morningstar2021density}, modifying objective functions \cite{zhang2020hybrid}, and exposing to OOD samples while training \cite{hendrycks2018deep}. Although such methods have promising results, they might require training and deploying a separate model in addition to the classifier, or rely on OOD data for training and/or hyper-parameter selection, which are not available in some applications. 
A Mahalanobis distance (MD) based OOD detection method \cite{lee2018simple} is a simpler approach which is easy to use. This method does not involve re-training the model and works out-of-the-box for any trained model. MD is a popular approach due to its simplicity.

Although MD based methods are highly effective in identifying \emph{far} OOD samples (samples which are semantically and stylistically very different from the in-distribution samples, e.g., CIFAR-10 vs. SVHN), we identify that it often fails for \emph{near} OOD samples (samples which are semantically similar to the in-distribution samples \cite{winkens2020contrastive}, e.g., CIFAR-100 vs. CIFAR-10) that are more challenging to detect. 
In this paper, we focus primarily on the near OOD detection task and investigate why the MD method fails in these cases. We propose \textit{relative Mahalanobis distance} (RMD), a simple fix to the MD, and demonstrate its effectiveness in multiple near-OOD tasks. Our solution is as simple to use as MD, and it does not involve any complicated re-training or training OOD data.
\section{Methods}
In this section, we briefly review the Mahalanobis distance method and introduce our proposed modifications to make it effective for near-OOD detection tasks.
\paragraph{Mahalanobis distance based OOD detection}
The Mahalanobis distance (MD)~\citep{lee2018simple} method uses intermediate feature maps of a trained deep neural network. The most common choice for the feature map is the output of the penultimate layer just before the classification layer. Let us indicate this feature map as $\vz_i=f(\vx_i)$ for an input $\vx_i$.
For an in-distribution dataset with $K$ unique classes, MD method fits $K$ class conditional Gaussian distributions $\gaussk, k=1, 2, \dots, K$ to each of the $K$ in-distribution classes based on the feature maps $\vz_i$.
We estimate the mean vectors and covariance matrix as: 
   $ \mathbf{\vmu}_k = \frac{1}{N_k} \sum_{i:y_i=k} \vz_i$, for $k=1, \dots, K,$
   and $\vSigma = \frac{1}{N}\sum_{k=1}^K \sum_{i:y_i=k}\left( \vz_i-\mathbf{\vmu}_k \right)(\z_i-\mathbf{\vmu}_k)^T$.
Note that class-conditional means $\mathbf{\vmu}_k$ are independent for each classes, while the covariance matrix $\Sigma$ is shared by all classes to avoid under-fitting errors. 

For a test input $\vx'$, the method computes the Mahalanobis distances from the feature map of a test input $\vz'=f(\vx')$ to each of the fitted $K$ in-distribution Gaussian distributions $\gaussk, k \in \{1,\dots,K\}$ given by $\text{MD}_k(\vz')$.  The minimum of the distances over all classes indicates the uncertainty score $\U(\vx')$ and its negative indicates the confidence score $\C(\vx') = - \U(\vx')$. These are computed as
\begin{align}
    \text{MD}_k(\vz') =& \left(\vz'-\mathbf{\vmu}_k\right)^T \vSigma^{-1} \left(\vz'-\mathbf{\vmu}_k \right), \\
     \C(\vx') 
     =& - \min_{k} \{ \text{MD}_k(\vz') \}.
     \label{eq:maha}
\end{align}
This confidence score is used as a signal to classify a test input $\vx'$ as an in-distribution or OOD sample.
\paragraph{Our proposed Relative Mahalanobis distance}
As we will demonstrate in Sec.~\ref{sec:failure:maha} and Appendix.~\ref{sec:simu}, OOD detection performance using MD degrades for near-OOD scenarios.
We draw our inspiration from the prior work by \citet{ren2019likelihood} showing that the raw density from deep generative models may fail at OOD detection and proposing to fix this using a likelihood ratio between two generative models (one modeling the sophisticated foreground distribution and the other modeling the background distribution) as confidence score.
Similarly, we propose Relative Mahalanobis distance (RMD) defined as
\begin{align}
\text{RMD}_k(\vz') \nonumber = \text{MD}_k(\vz')  - \text{MD}_0(\vz').
\end{align}
Here, $\text{MD}_0(\vz')$ indicates the Mahalanobis distance of sample $\vz'$ to a distribution fitted to the entire training data not considering the class labels: $\mathcal{N}(\mathbf{\vmu}_0, \vSigma_0)$, where $\vmu_0 = \frac{1}{N} \sum_{i=1}^N \z_i$ and $\vSigma_0 = \frac{1}{N}\sum_{i=1}^N \left( \vz_i-\vmu_0 \right)(\vz_i-\vmu_0)^T$.
This is a good proxy for the background distribution. %
The confidence score using RMD is given by
\begin{align}
    \C^{\text{RMD}}(\vx') &= -\min_k \{ \text{RMD}_k(\vz') \}.
    \label{eq:ratio_score}
\end{align}
See Appendix \ref{sec:algorithm} for the pseudocode.

RMD is equivalent to computing a likelihood ratio  $\max_k \left( \log p_k(\vz') - \log p_0(\vz') \right)$, where $p_k$ is a Gaussian fit using class-specific data  and $p_0$ is a Gaussian fit using data from all classes. 
Note that this can easily be extended to the case where $p_k$ and  $p_0$ are represented by more powerful generative models such as flows \cite{papamakarios2017masked, papamakarios2019normalizing}.  

Previous literature \cite{kamoi2020mahalanobis} discussed a similar topic however their work mainly focused on far-OOD, and their proposed method called Partial Mahalanobis distance (PMD) required a hyper-parameter (number of eigen-bases to consider), while our method performs better for near-OOD and is hyper-parameter free. 
See Appendix \ref{sec:partial_maha} for the comparison of PMD and RMD.

\section{Failure Modes of Mahalanobis distance}\label{sec:failure:maha}
\begin{figure}[h!]
    \centering
    \begin{subfigure}[b]{0.8\columnwidth}
         \centering
         \includegraphics[width=\textwidth]{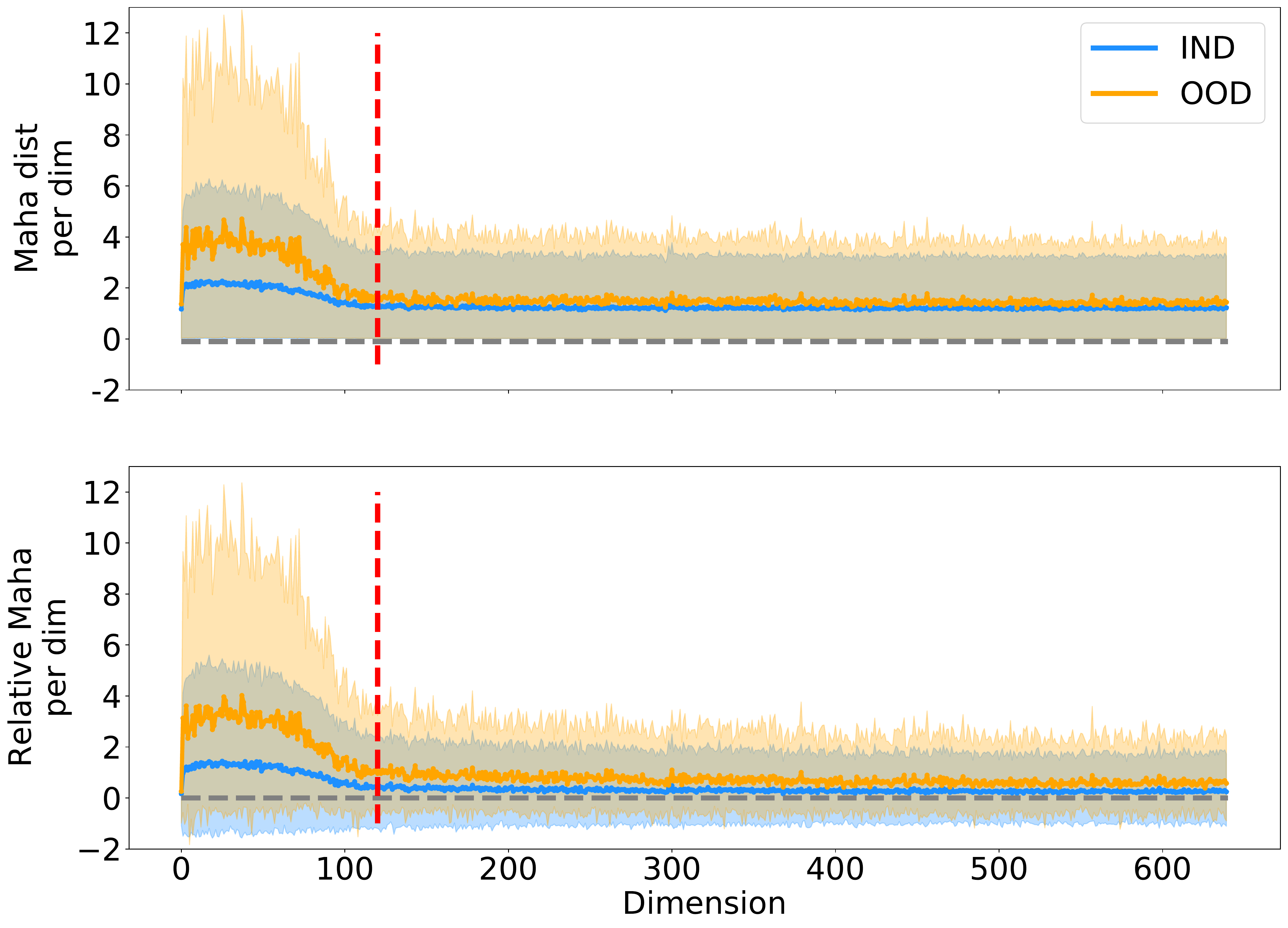}
         \caption{}
         \label{fig:eigen_cifar}
     \end{subfigure} 
     \vspace{-0.5em}
    \begin{subfigure}[b]{0.8\columnwidth}
        \centering
        \includegraphics[width=\columnwidth]{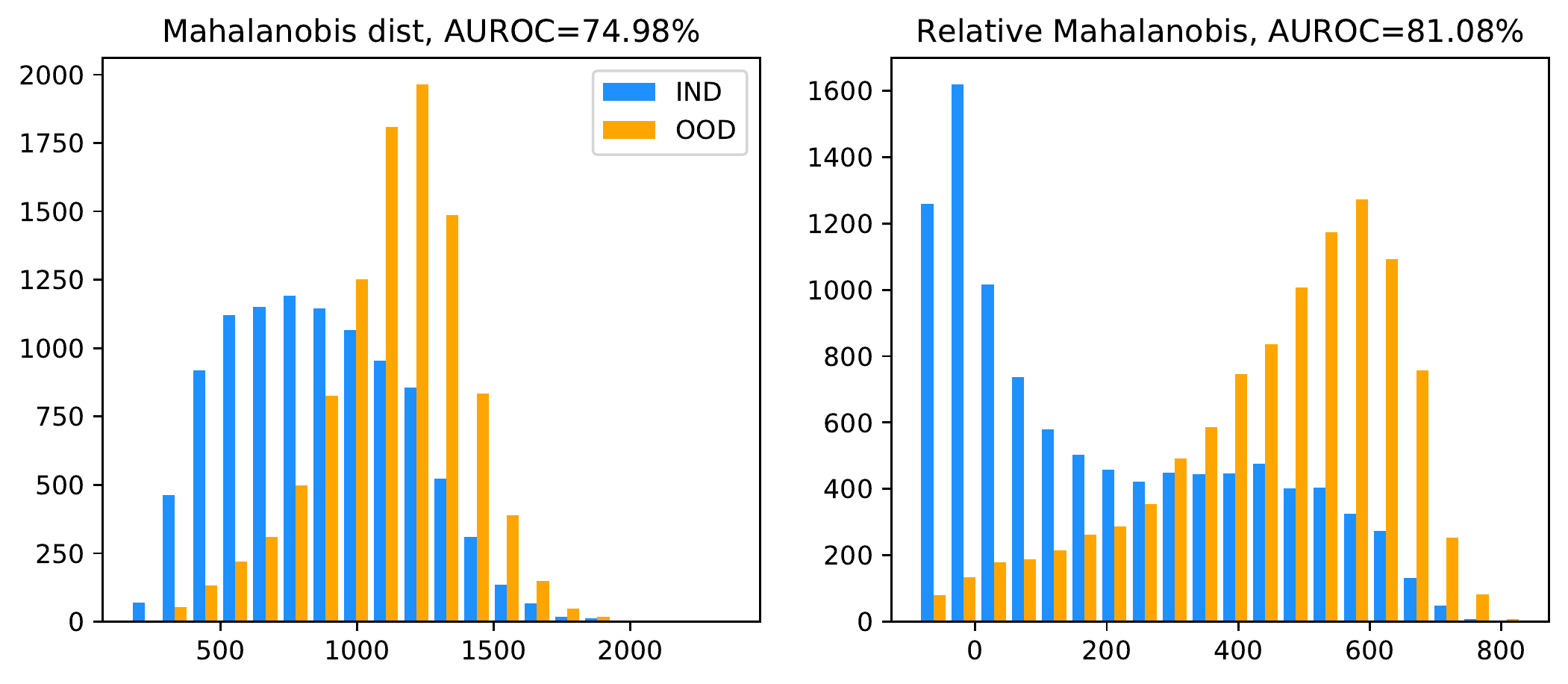}
        \caption{}
        \label{fig:hist_cifar}
     \end{subfigure} 
     \vspace{-0.6em}
    \caption{(a) Mahalanobis distance (top) and Relative Mahalanobis Distance (bottom) to CIFAR-100 (IND) and CIFAR-10 (OOD) along the $d^\mathrm{th}$ eigen-basis. %
    The solid lines represent the means over the IND and OOD test data respectively. The shading indicates the [10\%, 90\%] quantiles. The 120 top dimensions (before the red threshold) have distinct Mahalanobis distance between IND and OOD, while the later dimensions have similar Mahalanobis distances between IND and OOD, confounding the final score. 
    (b) Histograms of the Mahalanobis distance and Relative Mahalanobis distance for IND and OOD.}
\end{figure}
To better understand the failure mode of Mahalanobis distance and to visualize its difference from the Relative Mahalanobis, we perform an eigen-analysis to understand how these methods weight each dimension \citep{kamoi2020mahalanobis}. 
Specifically, we rewrite the Mahalanobis distance using eigenvectors $\vv_d$ of the covariance matrix $\vSigma$ as $\text{MD}(\vz') = (\vz' - \vmu)^T \vSigma^{-1} (\vz' - \vmu) = \sum_{d=1}^D l_d^2 / \lambda_d$, where $D$ is the dimension of the feature map, $\lambda_d$ is the $d^\mathrm{th}$ eigenvalue, and $l_d = \vv_d^T (\vz' - \vmu)$ is the projected coordinate of $(\vz' - \vmu)$ to the $d^\mathrm{th}$ eigen-basis $\vv_d$ such that $l_d^2 / \lambda_d$ can be regarded as the 1D Mahalanobis distance from the projected coordinate to the 1D Gaussian distribution $\mathcal{N}(0, \lambda_d)$. The $D$ eigen-bases are independent of each other.

 In the CIFAR-100 vs CIFAR-10 experiment, we found that OOD inputs have significantly greater mean distance (i.e. the average distance over the test samples) %
 in the top 120 dimensions with the largest eigenvalues, while in the remaining dimensions the OOD inputs have similar mean distance with the IND inputs (see Figure~\ref{fig:eigen_cifar}, top). 
Since the final Mahalanobis distance is the sum of the distance per dimension (this can be visualized as the area under the curve in Figure~\ref{fig:eigen_cifar}), we see that the later dimensions contribute a significant portion to the final score, overwhelming the top dimensions and making it harder to distinguish OOD from IND (AUROC=74.98\%).

Next we fit a class-independent 1D Gaussian as the background model in each dimension and compute RMD per dimension. As shown in Figure~\ref{fig:eigen_cifar} (bottom), using RMD, the contributions of the later dimensions are significantly reduced to nearly zero, while the top dimensions still provide a good distinction between IND and OOD.
As a result, the AUROC using RMD is improved to 81.08\%.

We conjecture that the first 120 dimensions are discriminative features that contain different semantic meanings for different IND classes and OOD, while the remaining dimensions are the common features shared by the IND and OOD. 
To support our conjecture,
we simulated a simple dataset following a high-dimensional Gaussian  with a diagonal covariance matrix and different means for different classes.
In particular, we set IND and OOD to have distinct means in the first dimension (discriminative feature) and the same mean in the remaining dimensions (non-discriminative features).
Since $\text{MD}$ is the sum over all the dimensions, the sum along those non-discriminative dimensions can overwhelm that of the discriminative dimension. 
As a result, the AUROC is only 83.13\%.
Using $\text{RMD}$, we remove the effect of the non-discriminative dimensions as for those dimensions the estimated $\mathcal{N}(\vmu_k, \vSigma) \approx \mathcal{N}(\vmu_0, \vSigma_0)$, detecting OOD perfectly with 100\% AUROC using the RMD.

\section{Experiments and Results}

As indicated in the previous section, in this work we primarily focus on near-OOD detection tasks. We choose the following established near-OOD setups: (i) CIFAR-100 vs. CIFAR-10, (ii) CIFAR-10 vs. CIFAR-100, (iii) Genomics OOD benchmark
\cite{ren2019likelihood} and (iv) CLINC Intent OOD benchmark \cite{larson2019evaluation, liu2020simple}.
As baselines, we compare our proposed RMD to traditional MD and maximum of softmax probability (MSP)~\citep{hendrycks2016baseline}, both working directly with out-of-the-box trained models. Note that most OOD detection methods require re-training of the models and complicated hyper-parameter tuning, which we do not consider for comparison.
We also ablate over different choices of model architectures with and without large scale pre-trained networks. The results are presented in the following sections.

\subsection{Models without pre-training}
In this section, we train our models from scratch using the in-distribution data. For CIFAR-10/100 tasks we use a Wide ResNet 28-10 architecture as the backbone. For genomics OOD benchmark we use a 1D CNN architecture consistent with~\citep{ren2019likelihood}.
For all benchmarks, at the end of training, we extract the feature maps for test IND and OOD inputs, and evaluate the OOD performance for our proposed RMD and comapre it with MD and MSP.
As seen in Table~\ref{tab:base_model}, contrasting MD and RMD, we observe a consistent improvement in AUROC for all benchmarks with gains ranging from 1.2 points to 15.8 points.
Comparing RMD to MSP, we observe a significant gain of 2.5 points for the Genomics OOD benchmark and partial gains for CIFAR-10/100 benchmarks.
This substantiates our claim that our proposed RMD boosts near-OOD detection performance.

\begin{table}[h!]
\resizebox{\columnwidth}{!}{
\begin{tabular}{llll}
\hline
Benchmark & MD & \makecell{RMD} & MSP \\ \hline
CIFAR-100 vs CIFAR-10 & 74.91\% & \textbf{81.01}\% & 80.14\%  \\ 
CIFAR-10 vs CIFAR-100 & 88.49\% & \textbf{89.71}\% &  89.27\% \\ \hline
Genomics OOD & 53.10\% \footnotemark & \textbf{68.98}\% & 66.53\% \\ \hline
\end{tabular}
}
\caption{Comparison of OOD-AUROC on the near-OOD benchmarks.}
\vspace{-0.5em}
\label{tab:base_model}
\end{table}

\vspace{-0.8em}
\paragraph{Using flows for $p_0$ and $p_k$}
To demonstrate that our proposed idea can be extended to more powerful density  models, we fit the feature maps using a one-layer masked auto-regressive flow \citep{papamakarios2017masked} for the CIFAR-100 vs CIFAR-10 benchmark. The AUROCs for using $\max_k \log p^{\mathrm{flow}}_k(\vz')$ and $\max_k (\log p^{\mathrm{flow}}_k(\vz') - \log p^{\mathrm{flow}}_0(\vz'))$ are 76.10\%, and 78.34\% respectively, showing that our proposal works for non-Gaussian density models as well. 

\footnotetext{We observed that the AUROC for MD changes a lot during training of the 1D CNN genomics model. We report the performance based on the model checkpoint at the end of the training without any hyperparameter tuning using validation set. See Section~\ref{sec:robust} for details.}

\vspace{-0.5em}
\subsection{Models with pre-training}
Massive models pre-trained on large scale datasets are becoming a standard practice in modern image recognition and language classification tasks. It has been shown that the high-quality features learnt during this pre-training stage can be very useful in boosting the performance of the downstream task \cite{hendrycks2019using, paul2021vision, fort2021exploring}.
In this section, we investigate if such high-quality representations also aid in better OOD detection and how our proposed RMD performs in such a setting, using different pre-trained models as architectural backbone for OOD detection. Specifically, we consider Vision Transformer (ViT) \citep{dosovitskiy2020image}, Big Transfer (BiT) \citep{kolesnikov2019big}, and CLIP \citep{radford2021learning} for CIFAR-10/100 benchmarks, and the unsupervised BERT style pre-training model \citep{devlin2018bert} for genomics\footnote{The BERT model used for the genomics benchmark is pre-trained on the genomics data with the standard masked language modeling method.} and CLINC benchmarks. 

We investigate two settings: (i) directly using pre-trained models for OOD detection and (ii) fine-tuning the pre-trained model on the in-distribution dataset for OOD detection.
\paragraph{Pre-trained models without fine-tuning}

We present our results in Table~\ref{tab:pretrain}, comparing MD and RMD for all benchmarks using different pre-trained models. Note that here we cannot evaluate MSP as the network was never trained to produce the predictive probabilities.
As shown, we first observe that, even without task-specific fine-tuning, the AUROC scores are either very close or better to Table~\ref{tab:base_model}, indicating that pre-trained models work well for OOD detection out of the box.
Secondly, we observe that RMD outperforms MD for all benchmarks with different pre-trained models with margins varying between 3.17 points to 16.5 points.
For the CIFAR-100 vs CIFAR-10 benchmark BiT models provide the best performance followed by CLIP and Vision Transformer. BiT with RMD achieves significantly higher AUROC (84.60\%) in comparison to the Wide ResNet baseline model (81.01\%).
For CIFAR-10 vs CIFAR-100, using pre-trained CLIP, RMD achieves 91.19\% AUROC, higher than any of the other methods considered.
Finally, it is worth noting that the gains provided by RMD are very prominent for genomics and CLINC intent benchmark when using BERT pre-trained features.

\begin{table}[h]
\centering
\resizebox{0.78\columnwidth}{!}{
\begin{tabular}{llll}
\hline
Benchmark & \makecell{MD} & \makecell{RMD}\\ \hline
\multicolumn{3}{c}{ViT-B$\_$16 Pre-trained} \\ \hline
CIFAR-100 vs CIFAR-10 & 67.19\% & \textbf{79.91}\% \\
CIFAR-10 vs CIFAR-100 & 84.88\% & \textbf{89.73}\% \\ \hline
\multicolumn{3}{c}{BiT R50x1 Pre-trained} \\ \hline
CIFAR-100 vs CIFAR-10 & 81.37\% & \textbf{84.60}\% \\
CIFAR-10 vs CIFAR-100 & 86.70\% & \textbf{89.87}\% \\ \hline
\multicolumn{3}{c}{CLIP Pre-trained} \\ \hline
CIFAR-100 vs CIFAR-10 & 71.40\% & \textbf{81.83}\% \\
CIFAR-10 vs CIFAR-100 & 83.57\% & \textbf{91.19}\% \\ \hline
\multicolumn{3}{c}{BERT Pretrained} \\ \hline
Genomics OOD & 48.46\% & \textbf{60.36}\% \\ \hline
CLINC Intent OOD & 75.48\% & \textbf{91.98}\% \\ \hline
\end{tabular}
}
\caption{Comparison of OOD-AUROC for the 4 near OOD benchmarks based on feature maps from pre-trained models. No fine-tuning involved. }
\label{tab:pretrain}
\end{table}

\vspace{-0.9em}
\paragraph{Pre-trained models with fine-tuning}
We now explicitly fine-tune the pre-trained model on the in-distribution dataset optimizing for classification accuracy.
Using the fine-tuned models for different benchmark, we report the performance in Table~\ref{tab:finetune}, comparing RMD with MD and MSP baselines.
We see that the performance of the MD improves significantly after the model fine-tuning (comparing Tables \ref{tab:pretrain} and \ref{tab:finetune}), suggesting a deletion of disruptive non-discriminative features which existed in the pre-trained models. MD achieves close or competitive AUROC when compared to RMD for most of the task evaluated, with the notable exception of genomics OOD (see Section \ref{sec:robust}).
In light of discussion in Section \ref{sec:failure:maha}, 
we conjecture that after task-specific fine-tuning using labeled data, most of the features become discriminative between IND and OOD. 
It is also possible that the pre-training and finetuning regimes end up at better local minima, and that the resulting features are capable of modelling the foreground and background implicitly (without our explicit normalization using RMD).
Therefore the effectiveness of RMD in such cases is limited.

\begin{table}[t]
 \centering
 \resizebox{0.95\columnwidth}{!}{
\begin{tabular}{llll}
Benchmark & \makecell{MD} & \makecell{RMD} & MSP \\ \hline
\multicolumn{4}{c}{ViT-B$\_$16 Fine-tuned} \\ \hline
CIFAR-100 vs CIFAR-10 & 
\textbf{94.42}\% & 93.09\% & 92.30\% \\
CIFAR-10 vs CIFAR-100 & \textbf{99.87}\% &
\textbf{98.82}\% & \textbf{99.50}\% \\ \hline
\multicolumn{4}{c}{BiT-M R50x1 Fine-tuned} \\ \hline
CIFAR-100 vs CIFAR-10 & 81.37\% & \textbf{84.60}\%  & 81.04\% \\
CIFAR-10 vs CIFAR-100 & \textbf{94.57}\% & \textbf{94.94}\% & 85.65\% \\ \hline
\multicolumn{4}{c}{BERT Fine-tuned} \\ \hline
Genomics OOD & 55.87\%\footnotemark & \textbf{72.04}\% & \textbf{72.02}\% \\ \hline 
CLINC Intent OOD & \textbf{97.92}\% & \textbf{97.62}\% & 96.99\% \\ \hline
\end{tabular}
}
\caption{AUROC for the 4 near OOD benchmarks based on feature maps from the fine-tuned models.}
\label{tab:finetune}
\end{table}

\footnotetext{We observed that the AUROC for MD changes a lot during finetuning. We report the performance based on the model checkpoint at the end of the training. See Section \ref{sec:robust} for details.}

\subsection{Relative Mahalanobis is more robust  %
}\label{sec:robust}

\begin{figure}[h!]
    \centering
    \begin{subfigure}[b]{0.4\textwidth} %
         \centering
         \includegraphics[width=\textwidth]{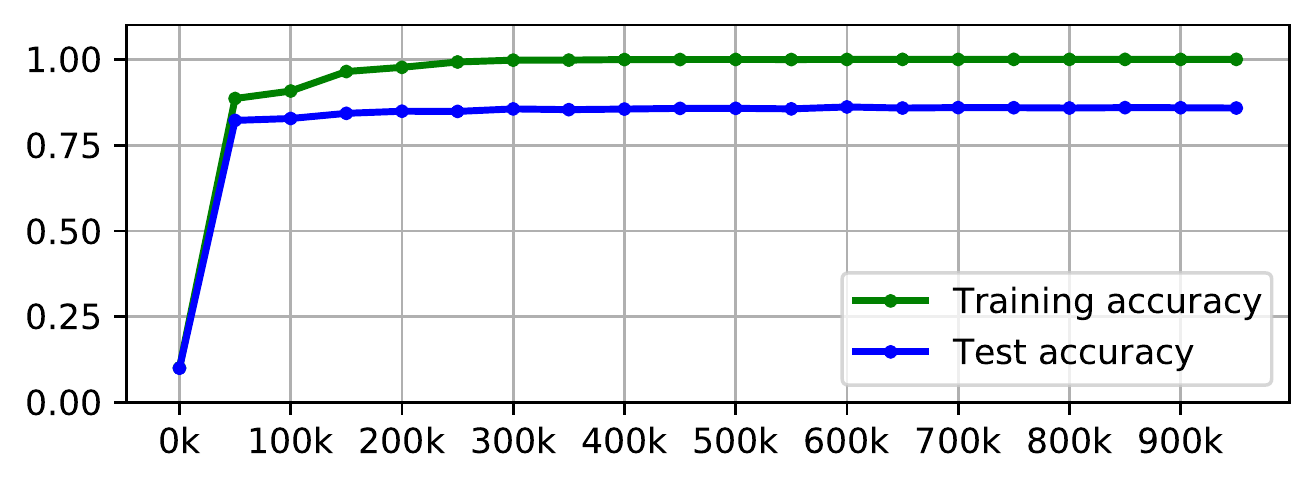}
         \caption{}
         \label{fig:train_vs_acc}
     \end{subfigure} 
     \vspace{-0.8em}
    \begin{subfigure}[b]{0.4\textwidth} %
    \centering
    \includegraphics[width=\textwidth]{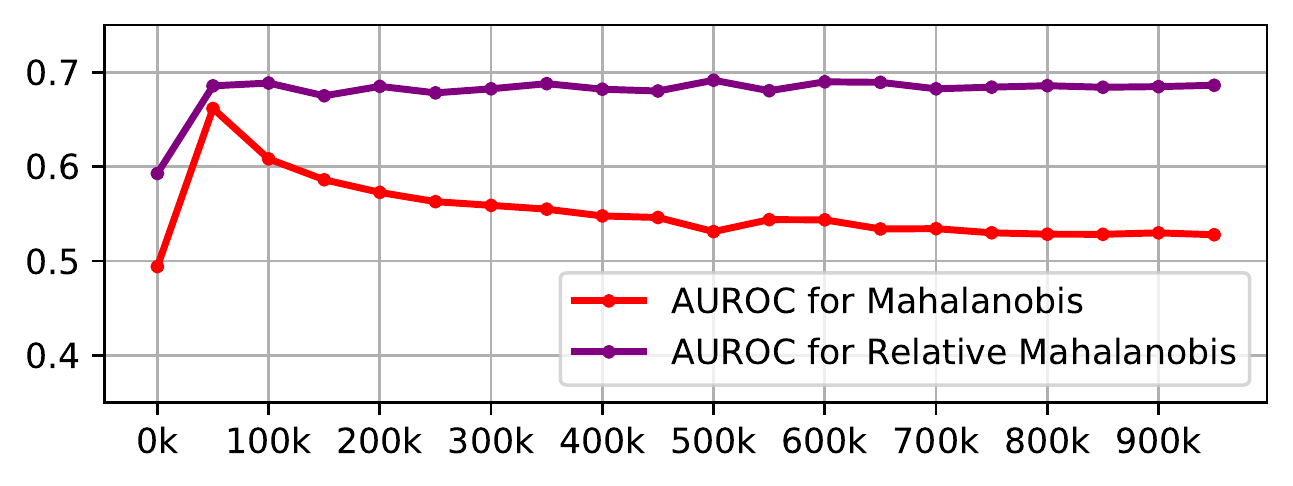} 
    \caption{}
    \label{fig:train_vs_auc}
     \end{subfigure} 
     \vspace{-0.1em}
    \caption{Comparison of MD and RMD as a function of training iterations:  MD performs well in the early stage of training, but drops significantly after that, while RMD stabilizes during training, which is consistent with the pattern of in-distribution accuracy.}
    \label{fig:training}
\end{figure}

In the genomics experiments, we noticed that the OOD performance of $\text{MD}$ is quite unstable during training of the 1D CNN model and the fine-tuning of the BERT pre-trained model. The AUROC of $\text{MD}$ increases at first during the early stages of training, followed by a decrease at later stages.
Figure~\ref{fig:training} shows the change of AUROCs for $\text{MD}$ and $\text{RMD}$ during the training of the 1D CNN model.
The AUROC of $\text{MD}$ quickly increases to 66.19\% at step 50k, when the model is not well trained yet, with training and test accuracies being 88.59\% and 82.20\% respectively. 	
As the model trains further and achieves higher training accuracy of 99.96\% and higher test accuracy of 85.71\% at step 500k, the AUROC for $\text{MD}$ drops to 53.10\%. 
On the other hand, the $\text{RMD}$ increases as the training and test accuracies increase, and gets stabilized as the accuracy stabilizes, which is a more desirable property to have. 
Similarly, we observed this phenomenon in the fine-tuning of the BERT genomics model. At the early training stage, AUROC for $\text{MD}$ achieves the peak of 77.49\%, while the model is not trained well with the training and test accuracies being only 82.62\% and 83.97\% respectively. 

\clearpage
\newpage
 \section*{Acknowledgements}
 We thank Zack Nado and D. Sculley for helpful feedback. 
 
\bibliographystyle{plainnat}
\bibliography{main}

\clearpage
\newpage
\appendix

\section{Pseudocode for Relative Mahalanobis distance} \label{sec:algorithm}

The pseudocode for our method is shown in Algorithm \ref{alg:rmd}.

\begin{algorithm}[H]
   \caption{Relative Mahalanobis distance}
   \label{alg:rmd}
\begin{algorithmic}[1]
   \STATE {\bfseries Input:} In-distribution train set $\mathcal{D}_{\mathrm{train}}^{\mathrm{in}} = \{ (\vz_i,y_i) \}$ with $K$ classes, in-distribution test set $\mathcal{D}_{\mathrm{test}}^{\mathrm{in}} = \{\vz'\}$, out-of-distribution test set $\mathcal{D}_{\mathrm{test}}^{\mathrm{out}} = \{\vz'\}$, feature extractor $\vz = f(\vx)$.
   \STATE Fit the $K$ class conditional Gaussian $\mathcal{N}(\vmu_k, \vSigma)$ using $\mathcal{D}_{\mathrm{train}}^{\mathrm{in}}$, where
   $ \mathbf{\vmu}_k = \frac{1}{N_k} \sum_{i:y_i=k} \vz_i$, for $k=1, \dots, K,$
   and $\vSigma = \frac{1}{N}\sum_{k=1}^K \sum_{i:y_i=k}\left( \vz_i-\mathbf{\vmu}_k \right)(\z_i-\mathbf{\vmu}_k)^T$.
   \STATE Fit the background Gaussian $\mathcal{N}(\vmu_0, \vSigma_0)$ using $\mathcal{D}_{\mathrm{train}}^{\mathrm{in}}$ ignoring the class labels, where $\vmu_0 = \frac{1}{N} \sum_{i=1}^N \z_i$ and $\vSigma_0 = \frac{1}{N}\sum_{i=1}^N \left( \vz_i-\vmu_0 \right)(\vz_i-\vmu_0)^T$.
   \STATE Compute $\text{MD}_k(\vz')$ based on $\mathcal{N}(\vmu_k, \vSigma)$, for each $\vz' \in \mathcal{D}_{\mathrm{test}}^{\mathrm{in}}$ and $\mathcal{D}_{\mathrm{test}}^{\mathrm{out}}$ using Eq. \ref{eq:maha}.
   \STATE Compute $\text{MD}_0(\vz')$ based on $\mathcal{N}(\vmu_0, \vSigma_0)$, for each $\vz' \in \mathcal{D}_{\mathrm{test}}^{\mathrm{in}}$ and $\mathcal{D}_{\mathrm{test}}^{\mathrm{out}}$.
   \STATE Compute RMD confidence score $-\min_k \{ \text{MD}_k(\vz') - \text{MD}_0(\vz') \}$ for each $\vz' \in \mathcal{D}_{\mathrm{test}}^{\mathrm{in}}$ and $\mathcal{D}_{\mathrm{test}}^{\mathrm{out}}$.
   \STATE Compute AUROC between $\mathcal{D}_{\mathrm{test}}^{\mathrm{in}}$ and $\mathcal{D}_{\mathrm{test}}^{\mathrm{out}}$ based on their RMD scores.
\end{algorithmic}
\end{algorithm}

\section{Additional Experimental Details}%

For CIFAR-10/100 experiments, we first train a Wide ResNet 28-10 model\footnote{\url{https://github.com/google/uncertainty-baselines/blob/master/baselines/cifar/deterministic.py}} from scratch using the in-distribution data. 
Next we use the publicly available pre-trained models ViT-B$\_$16\footnote{\url{https://github.com/google-research/vision_transformer}}, BiT R50x1\footnote{\url{https://github.com/google-research/big\_transfer}}, and CLIP\footnote{\url{https://github.com/openai/CLIP}},
and replace the last layer with a classification head and fine-tune the full models using in-distribution data. 
We do not fine-tune CLIP model since CLIP requires paired (text, image) data for training.
The fine-tuned ViT model has in-distribution test accuracy of 89.91\% for CIFAR-100, and 97.48\% for CIFAR-10.
The fine-tuned BiT model has in-distribution test accuracy of 86.89\% for CIFAR-100, and 97.66\% for CIFAR-10. 

For the genomics OOD benchmark, the dataset is available at Tensorflow Datasets\footnote{\url{https://www.tensorflow.org/datasets/catalog/genomics_ood}}.
The dataset contains 10 in-distribution bacteria classes, and 60 OOD classes and the input is a fixed length sequence of 250 base pairs composed by letters A, C, G and T. 
We first train a 1D CNN of 2000 filters of length 20 from scratch using the in-distribution data. 
We train the model for 1 million steps using the learning rate of $10^{-4}$ and Adam optimizer. 
Next we pre-train a BERT style model by randomly masking the input token and predict the masked token using the output of the transformer encoder. The model is trained using the unlabeled training and validation data. The prediction accuracy for the masked token is 48.35\%. 
At the fine-tuning stage, the model is fine-tuned using the in-distribution training data for 100,000 steps at the learning rate of $10^{-4}$, and the classification accuracy is 89.84\%. 

For CLINC Intent OOD, we use a standard BERT pretrained model\footnote{\url{https://github.com/google/uncertainty-baselines/blob/master/baselines/clinc_intent/deterministic.py}}, and fine-tune it using the in-distribution CLINC data for 3 epochs with the learning rate of $10^{-4}$. The classification accuracy is 96.53\%.

\section{Performance of Partial Mahalanobis distance} \label{sec:partial_maha}

\begin{figure}[h!]
    \centering
    \vspace{1em}
    \begin{subfigure}[b]{0.45\textwidth}
         \centering
         \includegraphics[width=\textwidth]{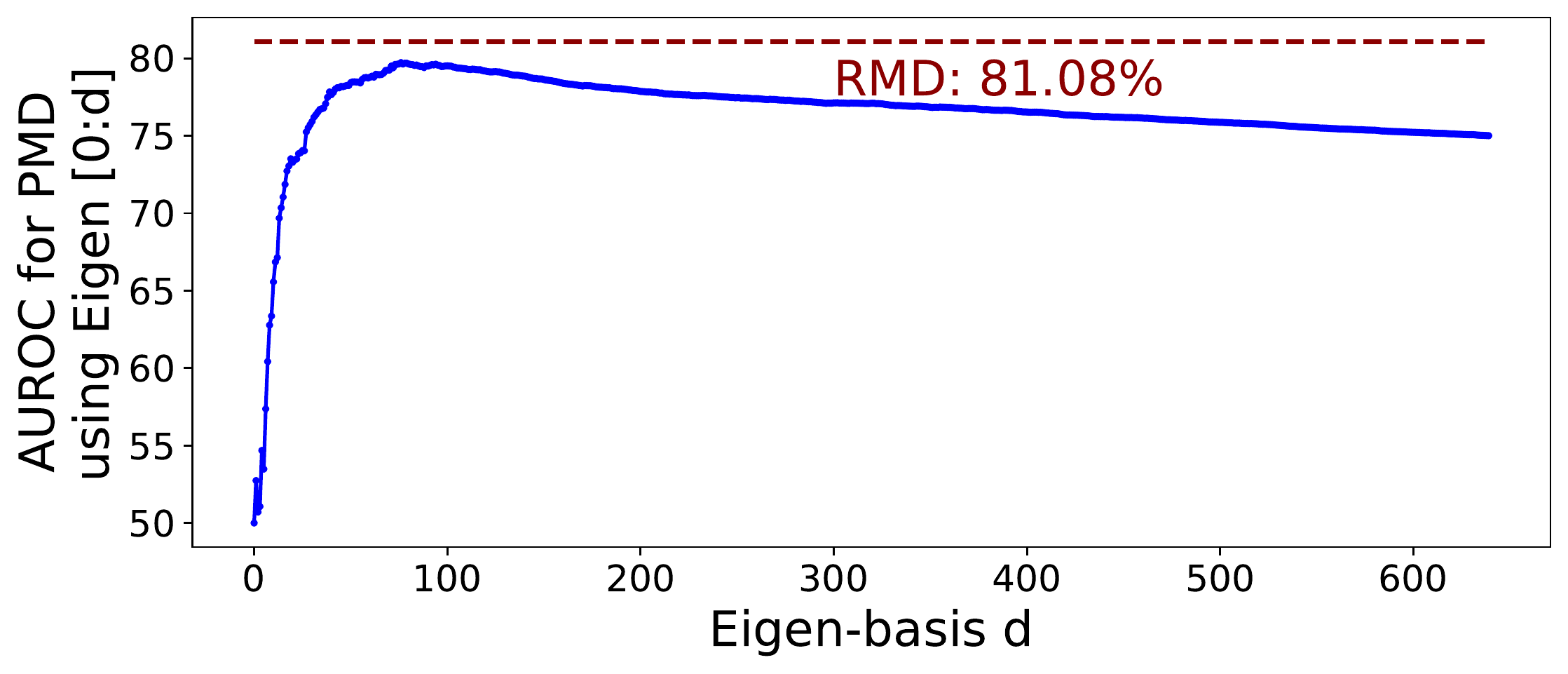}
         \caption{}
         \label{fig:pmd_largest}
     \end{subfigure} 
    \begin{subfigure}[b]{0.45\textwidth}
    \centering
    \includegraphics[width=\textwidth]{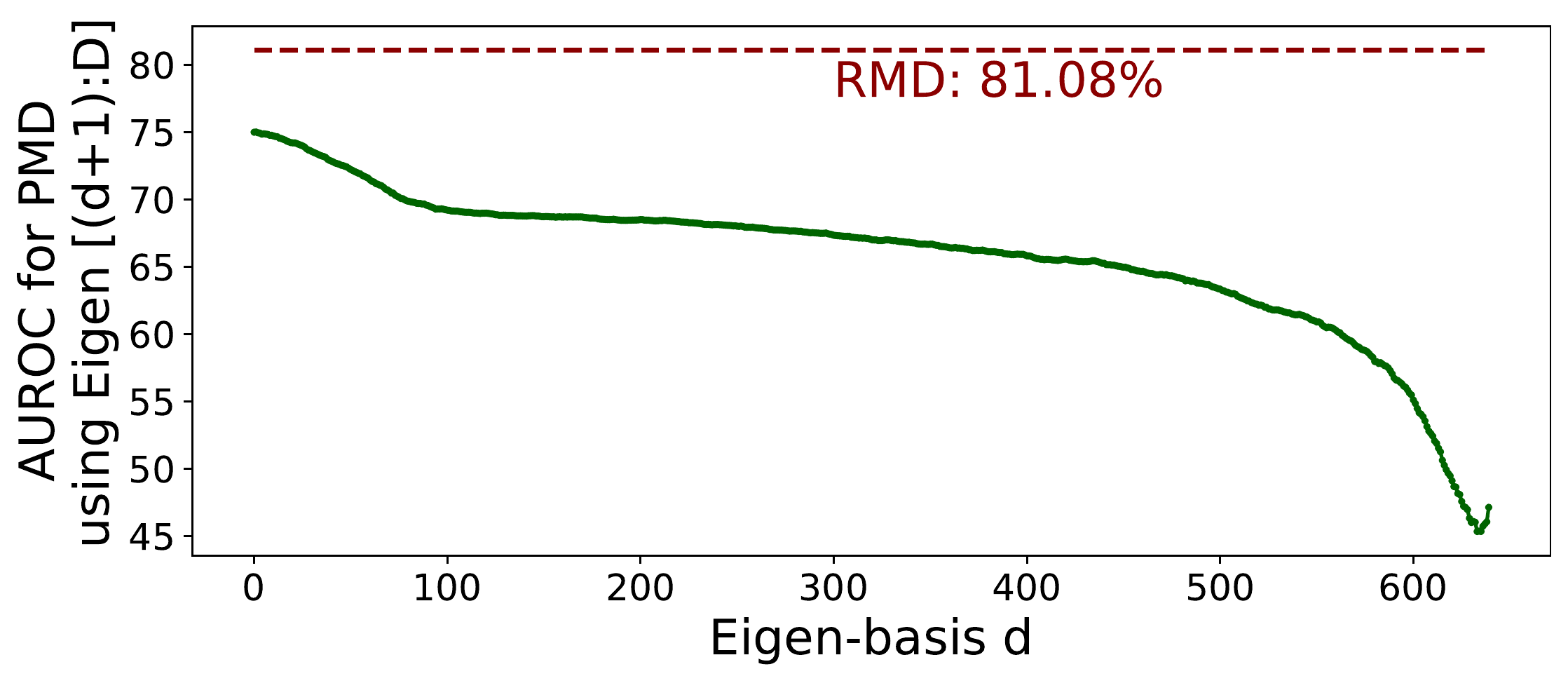}
    \caption{}
    \label{fig:pmd_smallest}
     \end{subfigure} 
    \caption{AUROC for Partial Mahalanobis distance (PMD) proposed in \cite{kamoi2020mahalanobis}. (a) PMD based on the first $[1:d]$ eigen-bases corresponding to the $d$ largest eigenvalues. (b) PMD based on the last $[d:D]$ eigen-bases corresponding to the smallest eigenvalues. The horizontal line indicates the AUROC for our method RMD.}
\end{figure}

We compare our method with the Partial Mahalanobis distance (PMD) proposed in \cite{kamoi2020mahalanobis}. 
PMD uses a subset of eigen-bases to compute the distance score, $\text{PMD}_S(\vz') = \sum_{d \in S} l_d^2 / \lambda_d, S \subset \{1, 2, \dots, D\}$. 
Although $S$ can be any subset of $\{1, \dots, N\}$, it was recommended to use $S = \{1,...,d\}$ or $S = \{d + 1,...,D\}$ corresponding to the largest or smallest Eigenvalues respectively.
We compare our RMD method with the two versions of PMD using the benchmark task of CIFAR-100 vs CIFAR-10. Since there is a hyperparameter $d$ involved in PMD, we search from $d=1, \dots, D$. 
Figure~\ref{fig:pmd_largest} shows the AUROC when using the top eigen-bases to compute PMD. 
The AUROC increases as $d$ increases and reaches to the peak of 79.72\% at $d=76$, and then decreases when including more dimensions.
Therefore the performance of PMD method depends on the choice of $d$, while our method RMD is hyperparameter-free. 
Our method also achieves a slightly higher AUROC of 81.08\% than the peak value for PMD.

We also investigate the performance of PMD when using eigen-bases corresponding to the smallest eigen-values (Figure~\ref{fig:pmd_smallest}). 
The AUROC decreases as we exclude the top eigen-bases from the set, suggesting that the top eigen-bases are more important for the near-OOD detection.
This observation supports our conjecture in Section \ref{sec:failure:maha} that the top eigen-bases are discriminative features and the rest are common features shared by the IND and OOD.

Another variant of the Mahalanobis distance called Marginal Mahalanobis distance (MMD) was also proposed in \cite{kamoi2020mahalanobis}. It fits a single Gaussian distribution to all the training data ignoring class, the same as we define the background model $p_0$ in our RMD. Though it has a good performance for far-OOD tasks (e.g. CIFAR-10 vs SVHN) \cite{kamoi2020mahalanobis}, it does not perform well for the near-OOD tasks, with AUROC being only 52.88\% for CIFAR-100 vs CIFAR-10, and 83.81\% for CIFAR-10 vs CIFAR-100.

\section{Simulation study for the failure mode of Mahalanobis distance} \label{sec:simu}

We use a simple simulation to demonstrate the failure mode of Mahalanobis distance. We simulate a binary classification problem where the two classes follow a high-dimensional Gaussian distribution with different means.
Specifically, $\vx \sim \mathcal{N}\left([a, 0, \dots, 0]_{1 \times D}, \sigma^2 I_{D\times D}
\right)$, where the covariance matrix is a fixed diagonal matrix with the scalar $\sigma$. The mean vector has only the first dimension non-zero. To distinguish the two classes, we set $a=-1$ for the first class, $a=1$ for the second class, and $\sigma=0.25$.
The key idea is that only the first dimension is a discriminative feature that is class-specific, whereas the remaining dimensions are non-discriminative common features that are shared by all classes. We set the number of dimensions $D=1024$. 
To simplify the problem, we set the covariance matrix to be diagonal such that the feature dimensions are independent.

\begin{figure*}[ht]
     \centering
     \begin{subfigure}[b]{0.24\textwidth}
         \centering
         \includegraphics[width=\textwidth]{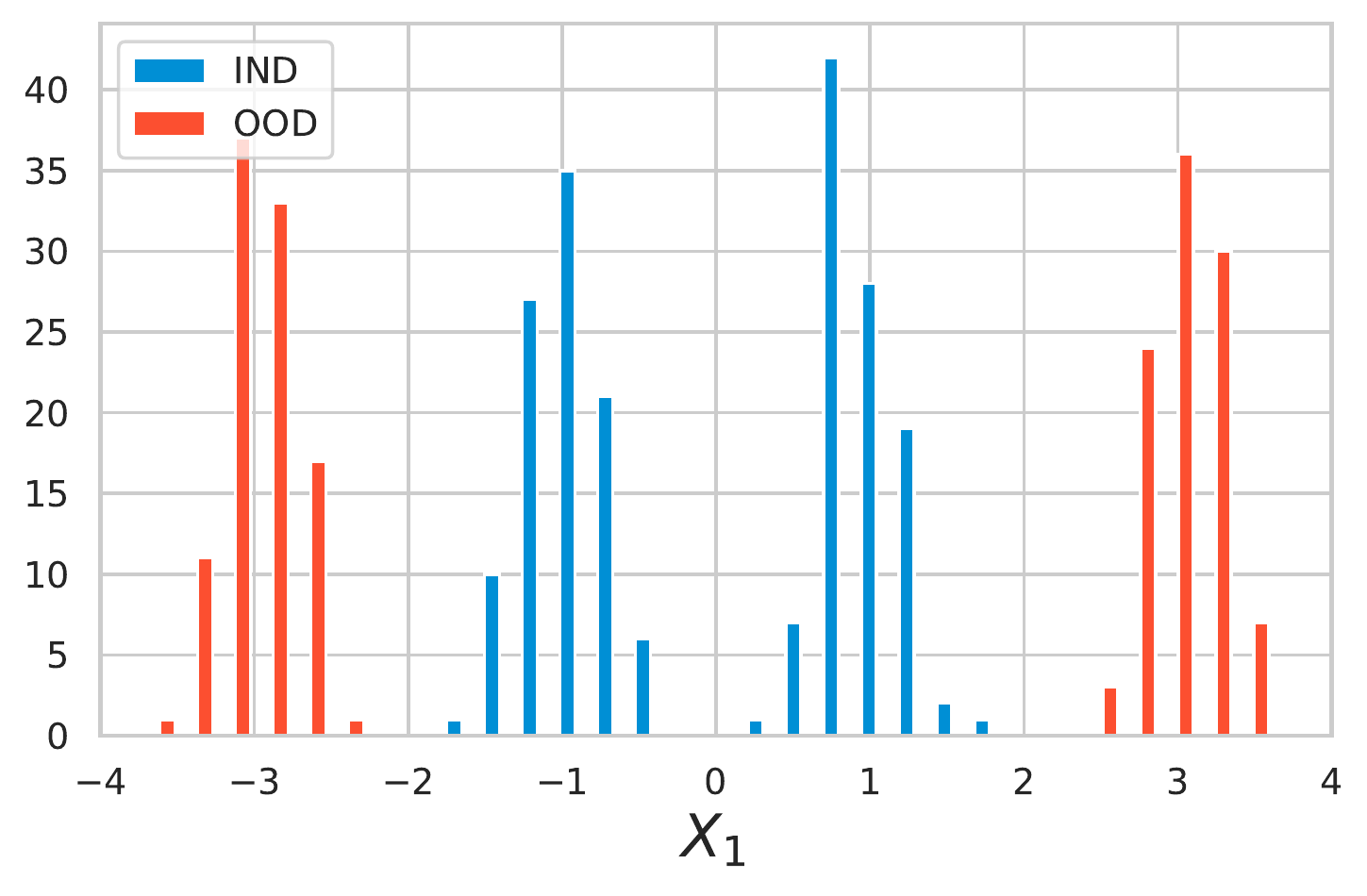}
         \caption{}
         \label{fig:toy_data_x1}
     \end{subfigure} 
    \begin{subfigure}[b]{0.24\textwidth}
    \centering
    \includegraphics[width=\textwidth]{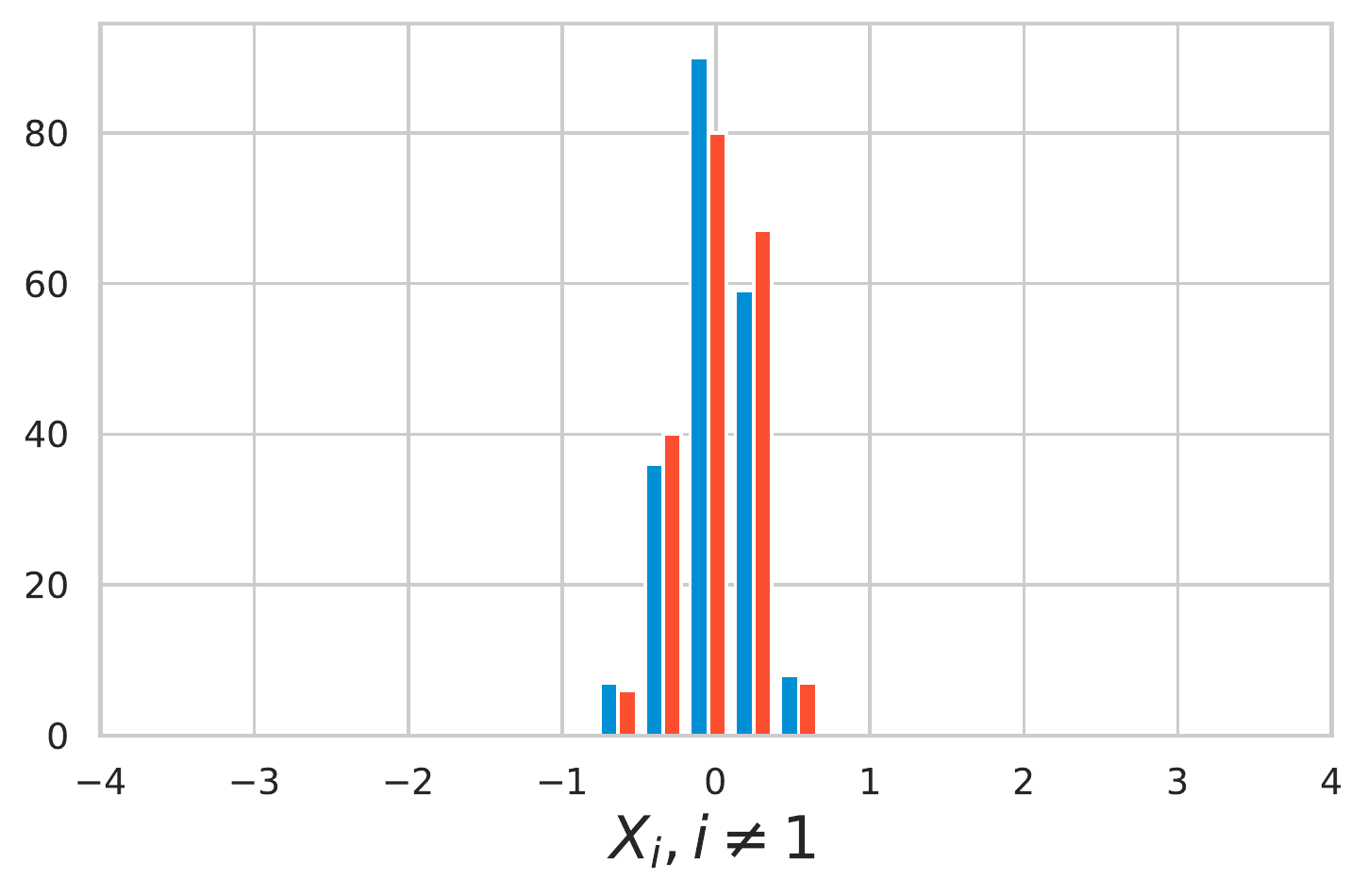}
    \caption{}
    \label{fig:toy_data_x2}
     \end{subfigure} 
     \begin{subfigure}[b]{0.24\textwidth}
         \centering
         \includegraphics[width=\textwidth]{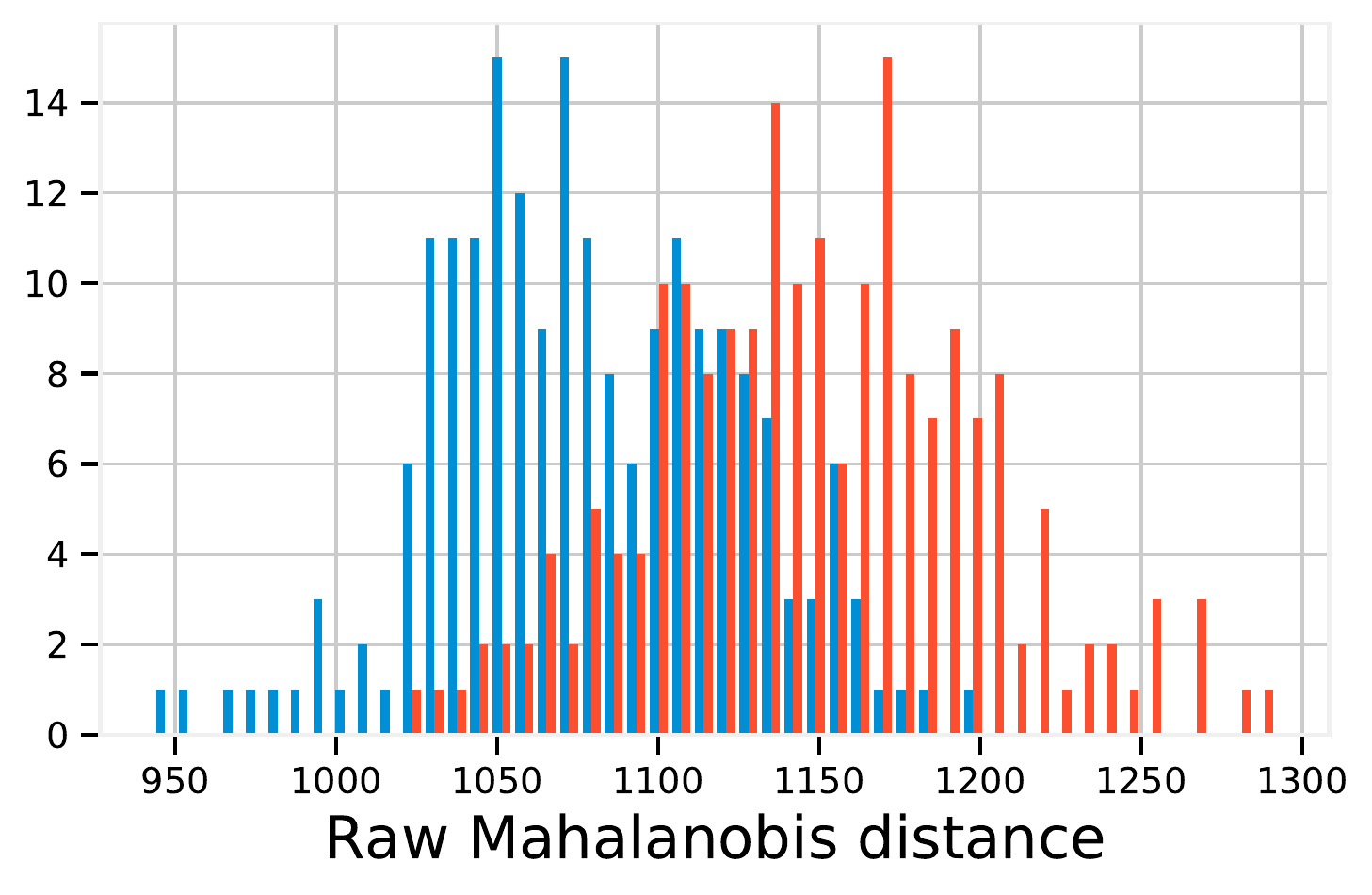}
         \caption{}
         \label{fig:toy_raw_maha}
     \end{subfigure}
     \begin{subfigure}[b]{0.24\textwidth}
     \centering
     \includegraphics[width=\textwidth]{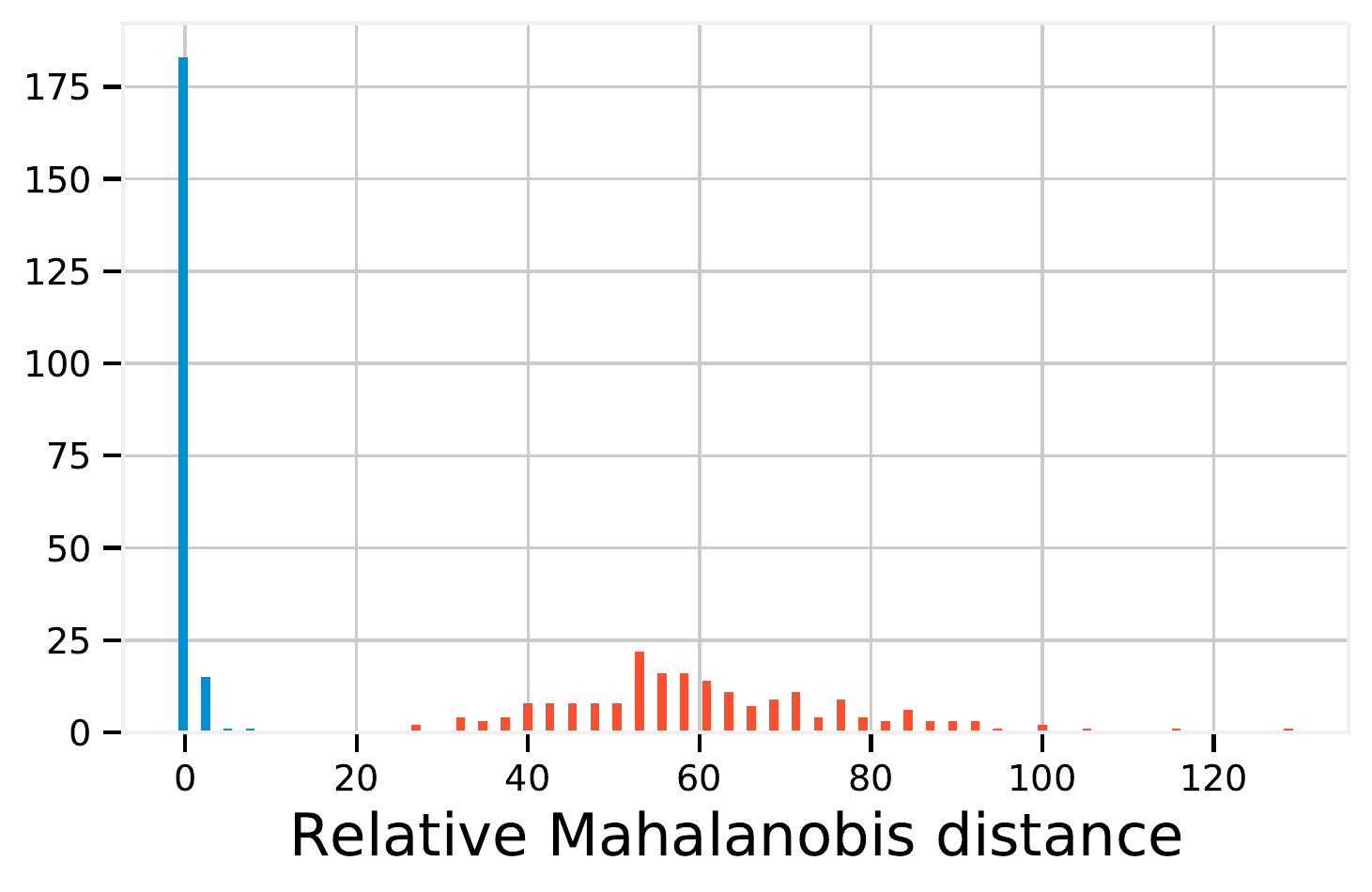}
     \caption{}
     \label{fig:toy_maha_ratio}
     \end{subfigure}
\caption{Simple simulation for the failure mode of Mahalanobis distance. (a) The histogram of the first dimension $x_1$ of IND and OOD data. The two IND classes (in blue) follow $\mathcal{N}(-1, \sigma^2)$ and $\mathcal{N}(1, \sigma^2)$ respectively, and the two OOD classes (in red) follow $\mathcal{N}(-3, \sigma^2)$ and $\mathcal{N}(3, \sigma^2)$ respectively. $\sigma=0.25$. (b) The histogram of $x_i, i \neq 1$. Both IND and OOD follow the same distribution $\mathcal{N}(0, \sigma^2)$. (c) The distributions of MD for IND and OOD inputs. The two distributions largely overlap. (d) The distributions of RMD for IND and OOD. The two distributions are well separated. OOD have positive values, while IND concentrate around zero.}
\label{fig:toy}
\end{figure*}

For each of the classes, we randomly sample $n=10,000$ data points from the given distribution for training data. For test data, we sample $100$ data points from each class as the test IND data. For test OOD data, we set $a=-3$ and $a=+3$ and sample $100$ data points from each of them. 
Figure \ref{fig:toy_data_x1} shows the histograms of the first dimension $x_1$ of IND and OOD data. The IND and OOD data points are well separated by the first dimension feature. Figure \ref{fig:toy_data_x2} shows the histogram of the remaining dimensions $x_i, i\neq 1$. The IND and OOD data points are not separable there, since they follow the Gaussian distribution with the same mean.   

For simplicity, we first treat the $\vx$ as the feature map $\vz$. We fit a class-conditional Gaussian $\mathcal{N}_k$ using the training data, and compute the MD for each of the test inputs. 
We find that although OOD inputs in general have a greater distance than IND inputs, the two are largely overlapping. See Figure \ref{fig:toy_raw_maha} for details. 

The reason behind the failure mode is simple. 
Since the dimensions are independent, the log-likelihood of an input is the sum of the log-likelihoods of each individual dimension, i.e. $\log p_k(\vx) = \log p_k(x_1) + \sum_{i\neq1} \log p_k(x_i)$, $k=1,\dots, K$. 
For the discriminative feature $x_1$, the distributions of IND and OOD are different, so approximately $\max_k \{\log p_k(x_1^{\mathrm{IND}})\} > \max_k \{\log p_k(x_1^{\mathrm{OOD}})\}$. 
However, the remaining non-discriminative features $x_i, i\neq1$ are class-independent and both IND and OOD inputs follow the same distribution. Thus the likelihood of IND inputs based on those features will be indistinguishable from that of OOD inputs, i.e. $\max_k \{\log p_k(x_i^{\mathrm{IND}})\} \approx \max_k \{ \log p_k(x_i^{\mathrm{OOD}})\}, i\neq1$.
When the number of non-discriminative features is much greater than the number of discriminative features, the log-likelihood of the former will overwhelm the latter. 

Next we compute the RMD. We fit a class-independent Gaussian distribution $\mathcal{N}_0$ using the training data regardless of the class labels, and compute the Relative Mahalanobis distance %
based on $\mathcal{N}_k, k=1,\cdots, K$ (class conditional Gaussian distribution) and $\mathcal{N}_0$ (class independent Gaussian distribution) for each of the test inputs. Using our proposed method, we are able to perfectly separate IND and OOD test inputs. See Figure \ref{fig:toy_maha_ratio}.

The class independent Gaussian $\mathcal{N}_0$ helps to remove the effect of the non-discriminative features. 
Specifically, since those non-discriminative features are class independent, the fitted class conditional Gaussian is close to the fitted class independent Gaussian, i.e. $\log p_k(x_i) \approx \log p_0(x_i), i\neq1$. 
Thus the two values are canceled by each other in the RMD computation, resulting in $\max_k \{\log p_k(\vx)\} - \log p_0(\vx) \approx \max_k \{\log p_k(x_1)\} - \log p_0(x_1)$. 
For the discriminative feature, the fitted class conditional Gaussian is very different from the fitted class independent Gaussian.
For IND inputs, $\max_k \{\log p_k(x^{\mathrm{IND}}_1)\} - \log p_0(x^{\mathrm{IND}}_1) > 0$, since the class conditional Gaussian fits better to the IND data.
For the OOD input, the difference between the two is nearly $0$, since none of the two distributions fit OOD. 
Therefore RMD provides a better separation between IND and OOD as we have seen in Figure \ref{fig:toy_maha_ratio}.

To mimic the real scenario where the feature maps are the extracted features from the neural networks, we train simple one-layer neural networks for this binary classification task. 
We retrieve the feature maps of the training data, fit a class conditional Gaussian and compute MD for the test inputs. 
We observed the same failure mode for this case; the distributions of MD between IND and OOD largely overlap. 
Then we fit a class independent Gaussian and compute RMD. Using RMD, we again recover the perfect separation between the two. 
We expect that the intermediate layer for image, text, and genomics models also contain non-discriminative features. Therefore our proposed method is useful for overcoming this effect and improving the performance of near-OOD detection. 

\end{document}

%% file: defs.tex
\newcommand{\z}{\mathbf{z}}

\newcommand{\gaussk}{\mathcal{N}(\mathbf{\vmu}_k, \vSigma)}

\newcommand{\C}{\mathcal{C}}
\newcommand{\U}{\mathcal{U}}

\usepackage{amsmath,amsfonts,bm}

\def\eqref#1{equation~\ref{#1}}

\def\1{\bm{1}}

\def\vmu{{\bm{\mu}}}

\def\vv{{\bm{v}}}

\def\vx{{\bm{x}}}

\def\vz{{\bm{z}}}
\def\vmu{{\bm{\mu}}}
\def\vSigma{{\bm{\Sigma}}}

\DeclareMathAlphabet{\mathsfit}{\encodingdefault}{\sfdefault}{m}{sl}
\SetMathAlphabet{\mathsfit}{bold}{\encodingdefault}{\sfdefault}{bx}{n}

